\documentclass[11pt,reqno]{amsart}
\usepackage[cp1251]{inputenc}
\usepackage[english,russian]{babel}
\usepackage{amsmath}
\usepackage{amssymb}
\usepackage{amsfonts}
\usepackage{graphicx}
\usepackage{xcolor}
\usepackage[colorlinks]{hyperref}

\usepackage{algorithm}%
\usepackage{algorithmic}%
\usepackage{listings}%
\usepackage{titlesec}
\titlespacing{\section}{0pt}{2.5ex}{1.5ex}
\titlespacing{\subsection}{0pt}{1.5ex}{1ex}
\titlespacing{\subsubsection}{0pt}{1.5ex}{1ex}
\titleformat{\section}{\large\bfseries\centering}{\thesection}{1em}{}
\titleformat{\subsection}[runin]{\bfseries}{\thesubsection.}{0.5em}{}[.\mbox{\ }]
\titleformat{\subsubsection}[runin]{\bfseries}{\thesubsubsection.}{0.4em}{}[.\mbox{\ }]

\newcommand\orcidicon[1]{${}$\ \ \ }

\begin{document}
\renewcommand{\refname}{References}

\thispagestyle{empty}

\title[ A Restart Rule for Genetic Algorithms]{\large A Restart Rule for Genetic Algorithms Based on Schnabel Census\footnote{This is a preprint of 
the article accepted for publication in the journal Siberian Electronic Mathematical Reports \copyright~Sobolev Institute of Mathematics SB RAS, 2026).}}
\author[A.V. Eremeev, Yu.V. Zakharova]{{\bf \large \textsc{A.V. Eremeev}}\protect\orcidicon{0000-0001-5289-7874}
{\bf \large \textsc{Yu.V. Zakharova}}\protect\orcidicon{0000-0003-4791-7011}\\ \\
}%
\address{Antion Valentinovich Eremeev  
\newline\hphantom{iii} Novosibirsk State University, 
\newline\hphantom{iii} 1, Pirogova str., 
\newline\hphantom{iii} Novosibirsk, 630090, Russia
\newline\hphantom{iii} Sobolev Institute of Mathematics,
\newline\hphantom{iii} pr. Koptyuga, 4,
\newline\hphantom{iii} 630090, Novosibirsk, Russia,
}
\email{eremeev@ofim.oscsbras.ru}
\address{Yulia Viktorovna Zakharova  
\newline\hphantom{iii} Sobolev Institute of Mathematics,
\newline\hphantom{iii} pr. Koptyuga, 4,
\newline\hphantom{iii} 630090, Novosibirsk, Russia,
\newline\hphantom{iii} Novosibirsk State University, 
\newline\hphantom{iii} 1, Pirogova str., 
\newline\hphantom{iii} Novosibirsk, 630090, Russia
}%
\email{yzakharova@ofim.oscsbras.ru}%

\vspace{1cm}
\maketitle {\small
\begin{quote}
\noindent{\bf Abstract:} A  new adaptive restart rule for Genetic Algorithms~(GAs) is considered. The rule is based on the Schnabel Census
method, originally developed for statistical estimation of a size of
animal population. In this paper, the Schnabel Census method is applied as a heuristic to estimate the number of
different solutions that may be visited with positive probability, given the current distribution
of offspring. The rule consists in restarting a GA as soon as the maximum
likelihood estimate reaches the number of different solutions
observed at the recent iterations.

We demonstrate how the new restart rule can be applied in a
GA with steady-state replacement scheme, using three different combinatorial optimization problems as examples.
Computational experiments on well-known benchmarks  show a statistically significant advantage 
of the GAs with the new restarting rule over the original versions of GAs. The new restart rule
also tends to be superior to the well-known rule, which restarts an
algorithm when the current iteration number is twice the number of iterations till the current best incumbent was found.
\medskip

\noindent{\bf Keywords:} genetic algorithm, restart rule, maximum likelihood, combinatorial optimization.
 \end{quote}
}

\bigskip

\section{Introduction}
\label{sec:intro}

Genetic Algorithms~(GAs) are the randomized search heuristics based on
the biological analogy of selective breeding in nature, originating
from the work of J.~Holland~\cite{Holl75}. A GA manipulates with a
{\it population} of~$\lambda$ {\it individuals}, using the random operators that
model mutation and crossover in nature. 
Suppose that a GA is applied to an optimization problem with
a finite space of solutions~$D$ and an objective function~$f:D\to \mathbb{R}$ to be maximized or minimized.
In this paper, we assume that the individuals of GA population are tentative solutions in~$D$.
The search in GAs is guided by the values of the {\it
fitness} function $\Phi(x)= \phi(f(x))$, which have to be evaluated for all individuals~$x$ in the
current population~$\Pi^t$ on iteration~$t$. Here $\phi: \mathbb{R} \to \mathbb{R}$ is an increasing 
function in the case of maximization problem, or a decreasing function in the case of minimization problem. 
The individuals of the initial population~$\Pi^0$ 
are generated according to some a priori defined probability distribution.

A GA may be considered as a Markov chain  in a number of ways.
For example, the states of the chain may correspond to different
vectors of ${\lambda}$ solutions that constitute the population
$\Pi^t$, see e.g.~\cite{Gu}. 
Some GAs are characterized by reducible Markov chains, and the population of individuals 
can be ``trapped'' in an area of the search space, from which the global optimum can not be reached. 
Other GAs satisfy the ergodicity requirements~\cite{Rud98},
ensuring the global optimum is found
after a finite number of iterations with probability~1. 
Nevertheless, it was shown in~\cite{DEL21} that many elitist evolutionary algorithms, also with the ergodicity property, require exponential time on average to leave some areas associated with local optima before they reach the global optimum. A similar behaviour may be expected from the GAs with crossover. 

In order to increase the probability of finding an optimal solution,
it is a common practice to use multiple restarts of a GA. 
%
The choice of the iteration when an evolutionary algorithm is stopped and restarted again
(a restart rule) was considered in a number of
papers on single-objective problems~\cite{BN98,Hulin,Jansen1,Luke,FKQS18} as well as the multi-objective problems~\cite{MGBM09}.
In some works, like~\cite{BN98,Hulin}, the restart times are chosen {\em adaptively}, i.e. the decision to restart can be made at any iteration, based on the search history of the algorithm on the given problem instance.  Other (non-adaptive) restart rules, like~\cite{Jansen1,Luke,FKQS18}, are set {\em a priori}, and do not directly make use of the search history at a given time step.
Theoretical analysis of non-adaptive restart rules may be found in~\cite{FKQS18,Jansen1,L93}, but most of the theoretical results 
are proved asymptotically, assuming that the problem size tends to infinity. 

This paper is a journal extended version of the conference publication~\cite{Er19}, where the new restart rule was proposed and tested on a GA for the set covering problem. Here in addition to the set covering problem, the
performance of the new rule is evaluated on the basis of GAs for the asymmetric traveling salesman problem and total weighted tardiness problem. 
%
This rule is based on the Schnabel Census method, transferred into the computer science from the statistics, where it was originally developed for the estimation of a size of
animal population~\cite{Seber}. 
The Schnabel Census, originally developed for estimation of animal
population size, is not used to count individuals here (the population size is a known parameter of a GA, kept
constant throughout the run), but
to estimate the number of different solutions that may
be visited with positive probability in this distribution, i.e., to estimate the size of support of this distribution. 
%
To make the Schnabel Census method
applicable, we accept several simplifying assumptions, in particular, that during the latest iterations, the GA population
was generated according to the same distribution.

The new restart rule 
consists in
restarting the GA as soon as a maximum likelihood estimate reaches
the number of different solutions observed at the latest
iterations. The rationale behind this rule is that it stops the GA
when, most likely, there are no more non-visited solutions in the
area where the GA population spent the latest iterations. In such
a case there are two appropriate ways to continue the search: either to restart the GA with a random new generation, or to modify the 
distributions of  mutation and crossover operators in order to leave the explored area. In the present paper we consider the first approach. 
The latter approach has received attention in~\cite{DLMN17,RW23,RW24}.  

There are two major GA outlines in use: the {\em
elitist} GAs, which copy a certain number of the ``most promising''
individuals from the previous population to the next one, and the {\em
non-elitist} GAs, which generate all individuals of a new population
with the same probability distribution. The well-known
GA~\cite{Holl75} is an example of non-elitist GAs.  
The new restart rule is 
applicable in both types of GAs, although in the present paper
the rule is tested only with elitist steady-state GAs.


The paper has the following structure. 
In the next section, we briefly describe the general outline of genetic algorithm with steady state population update,
which applies to all algorithms considered in this paper. In Section~\ref{sec:rule}, we describe the Schnabel Census and its usage in the new restart rule.  Section~\ref{sec:SCP} briefly outlines how this restart rule has been incorporated into a GA for the set cover problem and what computational results were obtained for it. Two applications of the general GA scheme to optimization problems on permutations are described in Section~\ref{sec:GAs_on_permut}. The results of experiments with these GAs, using the new restart rule and other alternatives are described in
Section~\ref{sec:experiment}. Concluding remarks a given in Section~\ref{sec:concl}.


\section{The Genetic Algorithm with Steady State Population Update} \label{sec:GA}

The GA starts with a random initial population, and
proceeds from the current population to the next one applying
selection, crossover (recombination) and mutation.
The initial population includes solutions generated randomly or using constructive heuristics. 
A result of the GA is the best solution found w.r.t. the objective function (fitness function). The number of individuals~$\lambda$ remains constant during the search. In a steady-state GA, only one
offspring is generated at each iteration. 
Algorithm~\ref{EA:algo} outlines the GA with steady state replacement~\cite{Davis27}.
Here the choice of parent individuals is made by means of one of the standard selection operators, e.g the tournament selection or proportional selection, or even with the uniform distribution.  To keep the population size constant, it is necessary to choose the individuals to be replaced by the offspring.  

As in many practical GA applications, in Algorithm~\ref{EA:algo}, before an offspring is added into the population, it undergoes a local improvement procedure. This procedure may implement some local search algorithm (then a GA is usually called a {\em memetic algorithm}~\cite{Moscato}) or a greedy heuristic or may just keep a solution unchanged. 

\begin{algorithm}
\begin{algorithmic}
\item[1:]
 Construct the initial population $\Pi^0$, set $t:=0$. 
\item[2:]
Until a termination condition is satisfied, perform steps
\begin{enumerate}
\item[2.1] Select two parent solutions $\pi^1$ and $\pi^2$ from $\Pi^{t}$.
\item[2.2] Apply a mutation operator to each parent solution, obtaining solutions $x^1$ and $x^2$.
\item[2.3] Generate an offspring $x'$ by applying a crossover operator to $x^1$ and $x^2$.
\item[2.4] Perform local improvements on $x'$ if necessary.
\item[2.5] Replace the individual of lowest fitness in the population~$\Pi^t$ by~$x'$, denote the new population~$\Pi^{t+1}$, set $t:=t+1$. 
\end{enumerate}
\item[3:] Return the best incumbent solution as a result.
\end{algorithmic}
\caption{Hybrid Evolutionary Algorithm with Steady State Replacement Scheme \label{EA:algo}}
\end{algorithm}

\section{Restart Rule Based on Schnabel Census}\label{sec:rule}

%

%

The new restart rule is based on the Schnabel Census method, 
which was developed in biometrics for statistical
estimation of the size of animal populations~\cite{Schnabel,Seber}. According to this method,
one takes repeated samples of size~$n_0$  from a population and counts the number of
\emph{distinct} animals seen. Often it is assumed that the
probability of catching any particular animal is the same. The
sampled animals are marked, unless they were marked
previously, and returned back into the population. Then
a statistical estimate for the total number~$\nu$ of individuals in
population is computed on the basis of the total number of 
animals marked in all the samples.

This method, with the sample size $n_0=1,$ was adapted
in~\cite{ReevesI,RE04} to estimate the number of local optima in
combinatorial optimization problems on the basis of repeated local
search outcomes with random starting points. 

In general, an event of falling into a basin of attraction of a local optimum number~$i$, or the event of catching 
a particular animal number~$i$ in a biological population can be viewed as realization of an integer random variable~$\xi$ 
 and one can set up a task to compute a statistical estimate for the number~$\nu$ of outcomes which have a positive probability measure.
 Unfortunately, without a-priory information on how the probabilities of different outcomes are related to each other, this problem does not have a satisfactory solution, see e.g.~\cite{Luke}. The reason is that in the general case, if some  value~$i$ of the random variable~$\xi$ has a probability that tends to zero, then with high probability the time till the first observation of this value will tend to infinity. So for any statistical estimate of~$\nu$ there is a counter-example demonstrating its negative bias. However, assuming a particular relationship on probabilities of different outcomes, one can obtain the maximal likelihood
estimate or a confidence interval for~$\nu$. In particular, one may assume that all outcomes are equally likely~\cite{ReevesI,RE04}, making the so-called {\em isotropic assumption}, or fit a certain type of parametric distribution (exponential, gamma, lognormal etc.) as in~\cite{GK}. 
Non-parametric estimates, such as the bootstrap or the jackknife, may also be
employed~\cite{ER,Pledger05,Seber}. 

In what follows, we will apply the Schnabel Census method to
estimate the number~$\nu$ of different offspring solutions~$\xi$ that have a non-zero probability to be created in the 
current iteration of a GA.  Note that given a specific problem instance, a specific population~$\Pi^t$, and the settings of all tunable parameters of the GA (population size, mutation and crossover rate etc.), it is possible to compute exactly the transition probabilities for the offspring population. However, such analysis is computationally prohibitive for large scale instances, which motivates our choice to use a statistical estimate for~$\nu$.  

Let a parameter~$r$ define the length of the historical
period considered for statistical analysis in the restart rule. Given a
value of~$r$, we assume that during
the $r$ latest iterations, all new offspring in the GA obeyed the same
distribution (assumption~1). 

In what follows, we assume that in the latest $r$~iterations of a GA,
the observed sample consists of $r$ independent offspring solutions (assumption~2). Let us define the
random variable~$K$ as the number of \emph{distinct} solutions
in this sample. We will make a simplifying isotropic assumption that all
solutions, that may be generated in the current distribution,
have equal probabilities (assumption~3). Then, as it was noticed in~\cite{Craig},  for any fixed $\nu$ the 
value $K$ has the following distribution:
\[  \Pr\{K=k\} = \frac{\nu!}{(\nu-k)!}\frac{S(r,k)}{\nu^r} ,   \]
where $S(r,k)=\frac{1}{k!}  \sum_{s=0}^k (-1)^k {k \choose s}
(k-s)^r$ is the Stirling number of the second kind. This distribution is also known as the Arfwedson
distribution~\cite{JK}. The maximum
likelihood estimate~$\hat{\nu}^{\rm ML}$ for~$\nu$ is
\begin{equation}\label{eqn:nu_max}
\hat{\nu}(r,k)=\mbox{argmax} \left\{\frac{\nu!}{(\nu-k)!
\nu^r}\right\},
\end{equation}
where $k$ is the number of different solutions actually generated
on the latest $r$ iterations of the GA. The value $\hat{\nu}(r,k)\ge k$ may be found from~(\ref{eqn:nu_max}) by the
standard one-dimensional optimization methods (see
e.g.~\cite{ReevesI}).

The proposed rule restarts the GA as soon as the
estimate~$\hat{\nu}(r,k)$ becomes equal to~$k$. The value of~$r$ is 
tuned adaptively during the GA execution. The
rationale behind this rule is that once the inequality
${\hat{\nu}(r,k)=k}$ is satisfied, most likely there are no more
non-visited solutions in the area where the GA population spent
the latest~$r$ iterations. In such a situation, it is more appropriate
to restart the GA rather than to wait till the population
distribution will significantly change by the evolutionary mechanisms.

Clearly, there are situations where assumptions~1-3 may be irrelevant in practice, but when a GA spends a significant
number of iterations without any progress, assumptions 1 and 2 are likely to hold. Assumption~3 is not 
so likely, but in this work it is accepted for the reason of computational simplicity. We acknowledge that the isotropic assumption
might lead to negative bias of the estimate~$\hat{\nu}(r,k),$ and more adequate estimates of~$\nu$ would allow longer runs of the GA. 

We compare the new restart rule based on the Schnabel Census with two other alternatives, 
considering the following three modes of GA execution:
\begin{itemize}
\item Mode~A.~Single run with no restarts.

\item Mode~B. Restart the GA as
soon as the current iteration number becomes twice the iteration
number when the best incumbent was found. This adaptive restart rule was used
successfully by different authors to restart random
hill-climbing method~\cite{HK93} and GAs~\cite{BN98,EK17}.
To avoid early restarts, this rule is applied only after
a certain number of iterations, denoted by~$t_{\min}$. We use
$t_{\min}$ equal to the population size.

\item Mode~C. Restart the GA using the adaptive restart rule based on the Schnabel Census.
The value of parameter $r$ is chosen 
adaptively as follows: {\em Whenever the best
found solution is improved, $r$ is set to be the population size.
If the best incumbent was not improved during the latest $2r$
iterations, then the value of~$r$ is doubled.} We reset $r$ to the
population size when the best
found solution is improved, assuming that whenever the best incumbent is
improved, this means that the population has reached a new unexplored area and 
 the length of the historic period for analysis should be reduced. 
To reduce the CPU cost, the
termination condition is checked only when the
value of~$r$ is updated.
\end{itemize}

\section{Previous Results for Set Cover Problem}
\label{sec:SCP}

The Set Cover Problem may be
formulated as follows. {\it Consider a set ${\mathcal
M}=\{1,\ldots,m\}$ and the subsets ${\mathcal M}_j\subseteq
{\mathcal M}$, where $j\in {\mathcal N}=\{1,...,n\}$. A subset
${\mathcal J} \subseteq {\mathcal N}$ is a cover of ${\mathcal M}$
if $\bigcup_{j \in {\mathcal J}}{{\mathcal M}_j}={\mathcal M}$.
For each ${\mathcal M}_j$, a positive cost~$c_j$ is assigned. The
SCP is to find a cover of minimum summary cost.}


The SCP is a well-known NP-hard problem~\cite{GJ}. 
A number of heuristic algorithms have been
developed for approximate solving the SCP: Lagrangian relaxation heuristics~\cite{CFT}, neural
networks~\cite{GW97}, local search~\cite{YKI}, GAs~\cite{BC96,E99}, ant colony algorithms~\cite{AlKo}
etc. Here we consider the GA for the SCP which was
proposed initially without any restart rule in~\cite{E99}. This GA follows the general scheme of Algorithm~\ref{EA:algo} and it is denoted
as~NBGA because of the non-binary representation of
solutions~\cite{BC96,E99},  involving an alphabet with up to~$n$ symbols.
The NBGA uses a problem-specific
crossover operator based on the linear programming, the proportional selection operator and a problem-specific mutation
operator.
The offspring are improved by the means of three greedy
procedures before they are added into the population.

In~\cite{Er19} the new restart rule was incorporated into the NBGA. 
Computational experiments on benchmark instances
from OR-Library were aimed  at comparing three modes of GA execution (A,B and C) given equal computational
budget, measured in terms of GA iterations.


A single experiment with a GA, given a certain computational budget, will be called {\em a
trial}. In~\cite{Er19}, $N=30$~independent trials of the NBGA in each of the three
modes were carried out. 
In what follows, $F_{\rm bst}$ will denote the
frequency of obtaining a solution with the best known cost from
the literature, estimated by 30~trials.

A statistical analysis of experimental data was carried out using
the significance test from~\cite{BH1977} (see~Ch.~8,~\S2), comparing two algorithms in terms of probability of
finding an optimal or a best-known solution.

Comparison of the GA results in modes~A and~C on these problems showed  that among~37 instances, where these two modes yielded different frequencies~$F_{\rm bst}$, mode~C had a higher value~$F_{\rm bst}$ in 31~cases and in 16 out of these 31~cases the difference was statistically significant. Mode~A had a statistically
significant advantage to mode~C only on a single instance.

Modes~B and~C show different
frequencies~$F_{\rm bst}$ on 28 randomly generated instances. On 16 of these 28~instances,
mode~C had a higher value~$F_{\rm bst}$ than
mode~B and in 5 out of these 16~cases the difference was
statistically significant. Mode~B had a statistically
significant advantage to mode~C only on a single instance.

In terms of percentage of deviation from the best-known cost, which was averaged over all
instances of series {\it 4-6}, {\it a,c,d} and {\it e,f,g,h},
mode~C gave the least error.
Besides that, it was observed that the restart mode~C terminated the GA later than mode~B in
most of the cases. This is natural because the restart
rule of mode~B is based solely on the values of objective
function of the obtained solutions, while the new restart rule
uses the information about the generated covers and their
frequencies. In cases where mode~C
yielded statistically greater frequency of finding the optima,
compared to mode~B, further analysis included the average number of iterations~$t_{\rm
avg}$ that were made until the
restart rule terminated a GA 
(over 30 runs). This analysis showed that in mode~C the value of~$t_{\rm avg}$ was 4-8 times larger than in mode~B.

The SCP instances, where the frequency of finding optima was equal to~100\%, were considered in order to evaluate the overall CPU cost of the GA execution (including the CPU time for Schnambel Census
estimate~$\hat{\nu}(r,k)$). On average, in terms
of the CPU time and in terms of the total number of tentative
solutions made until first visiting an optimum, mode~C tended to be
more efficient than the other two modes.

The three modes of running the GA were also tested on two series
of combinatorial  unicost SCP instances from the OR-library, denoted~{\it clr} and {\it stein}. In these experiments,
restart mode~C showed better or equal results compared to the other two modes, except
for a single instance clr13 with the largest number of variables.
On clr13, the best known solution was found only in mode~A and it took more than 4000~iterations. Mode~C was inefficient on this instance, probably due to a negative bias of the maximum likelihood estimate~$\hat{\nu}(r,k)$ from formula~(\ref{eqn:nu_max}). 
The statistics~$t_{\rm avg}$ indicates that the restart rule of mode~B triggered too early on the combinatorial problems, precluding the GA from finding good solutions. We suppose that the negative bias of modes B and C on clr13 is more evident than in the case of randomly generated SCP instances because the combinatorial unicost instances tend to have larger plateaus of solutions with equal objective function values. 


\section{Genetic Algorithms with Restarts for Two Optimization Problems on Permutations}\label{sec:GAs_on_permut}
\subsection{The Genetic Algorithm for Traveling Salesman Problem}

 The {\em Traveling Salesman Problem (TSP)} is one of the classical problems of discrete optimization. 
Given a complete directed graph~$G$ 
with vertex set $V=\{v_1,\dots,v_n\}$ and arc set~$A=\{(v_i,v_j):\ v_i,v_j\in V,\ i\ne j\}$.
For each arc $(v_i,v_j)\in A$, a weight (arc length) ${c(v_i,v_j)\ge 0}$ is specified.
The objective is to find a Hamiltonian cycle $C$ (traveling salesman's tour) with the minimum total length 
$\sum_{(v_i,v_j)\in C}c(v_i,v_j)$.
If the length of arc $(v_i,v_j)$ differs from the length of the reverse arc $(v_j,v_i)$ for at least one pair of vertices~$v_i,\
v_j$, then the traveling salesman problem is called {\em asymmetric} (ATSP).
The asymmetric traveling salesman problem arises in various practical applications:
scheduling problems considering equipment changeover,
vehicle routing problems, shop scheduling problems, goods and services distribution, production planning, etc.

The asymmetric traveling salesman problem is strongly NP-hard~\cite{GJ},
which justifies the development of heuristic algorithms and metaheuristics, particularly genetic algorithms. Special-purpose metaheuristics were proposed for the ATSP: SAX/RAI Memetic Algorithm~\cite{BFM2004}, hybrid approach with position-based representation of solutions~\cite{XCYHSC2008}, GA with GAPX crossover~\cite{TWO14}, GA with EAX crossover~\cite{NS12}, Doubly-Rooted Stem-and-Cycle Ejection Chain Algorithm~\cite{RGG2016}, Variable Neighborhood Search~\cite{BCK2001} and others.

Our genetic algorithm uses the scheme of Algorithm~\ref{EA:algo}.
The initial population is constructed using the ``arbitrary insertion'' heuristic~\cite{YI} and the heuristic by W.~Zhang~\cite{Zhang00}. 
During population evolution, problem-specific mutation and optimal recombination~\cite{EremeevKovalenko2020} are employed.
Individuals in both initial and final populations are improved through local search heuristics based on 3-opt neighborhood.

The selection operator employs a tournament selection.
Two mutation operators are implemented, performing random descent in either 3-opt or 4-opt neighborhood.
The operators are used with equal probability.
The crossover operator solves the optimal recombination problem using the algorithm presented in~\cite{Kov}.

The balance between the intensity of local optimization and population-based search is regulated through adaptive restart based on the Schnabel census method (mode~C) or the classic restarting rule (mode~B).


\subsection{Genetic Algorithm for the Total Weighted Tardiness Problem}

We consider the following problem of scheduling a set of jobs $\mathcal{J}=\{1,\dots,n\}$ on a single machine.
Job $j\in \mathcal{J}$ is characterized by release date $r_j$, processing time $p_j$,  positive weight $w_j$, and due date $d_j$ by which it
should be completed. The machine can execute at most one job at a time
and preemptions are disallowed.
Let $C_j$ denote the completion time of job $j\in \mathcal{J}$,
then the tardiness $T_j$ of job~$j$ is computed as $\max\{0; C_j-d_j\}$.
The goal is to construct a schedule such
that the total weighted tardiness $\sum_{j\in \mathcal{J}} w_j T_j$ is minimized.
Using the classical three-field notation our problem is denoted as $1|r_j,d_j,w_j|\sum_j w_j T_j$
and called {\em the one-machine total weighted tardiness problem (TWTP)}.
This problem is NP-hard and arises in several practical
settings, in particular in the chemical industry~\cite{WDK2002}.
In addition, scheduling models with a single machine also have
implications for scheduling research involving multiple machines, where the
results obtained from single-machine problems can often be applied
to more complex scheduling environments, such as parallel machines, flow shops, job shops, and  open shops.

The state-of-the-art metaheuristics for the TWTP are genetic algorithm with local improvements (GA-LI) from~\cite{AAS2003},
population-based variable neighborhood search algorithm (PVNS)~\cite{WT2009},
multiple VNS algorithm (m-VNS)~\cite{CFLL2016} and
hybrid evolutionary algorithm (HEA-DS) from~\cite{DLCX2017}. 



Our GA uses the scheme of~Algorithm~\ref{EA:algo}. 
For the construction of the initial population we use the following main constructive heuristic:
a random non-scheduled job is selected at each step; the position is assigned to this job such that it gives the best objective for the current partial solution.
Also, the population contains permutations, that correspond non-decreasing order of due dates $d_j$, non-increasing order of $w_j/p_j$ and non-increasing order of $(w_j/p_j)\cdot \exp\{-\max \{0,(d_j-p_j)\}/(n\cdot p_{aver})\}$,
where $p_{aver}=\sum_{j=1}^n p_j/n$.  

Individuals of the last population is improved by a local optimization procedure based on the {\em swap} neighborhood (positions of two jobs are exchanged) and the {\em insert} neighborhood (a job is inserted in some other position). 
For reducing the running time of local search heuristics, we use strategies provided  in~\cite{Geiger2010}  for  $1|r_j=0,d_j,w_j|\sum_j w_jT_j$. 
In particular,  obviously unpromising moves are excluded and only a part of the neighborhood is considered by moving jobs only from positions
located at the distance of no more than $20\% n$ from each other.

We use two mutation operators that
perform a random jump within swap or insert neighborhood~\cite{Geiger2010}.
The operators are used for mutation with equal probability. The mutation is applied with probability $p_\mathrm{mut}$.
In the crossover operator we solve the optimal recombination problem by  method from \cite{Kov} with probability $p_\mathrm{cross}$ (this crossover is called {\em OCX}, Optimized Cycle Crossover), and use the well-known {\em OX} (Ordered Crossover)~\cite{Kellegoz2008} with probability $1-p_\mathrm{cross}$.


The preliminary version of our GA and investigation of various crossover operators were presented at the International conference MOTOR-2023~\cite{Zakharova23}, but restarting rules were not analysed. In the current research GA will be tested in three modes A, B and C as in the case of ATSP.

\section{Computational Experiments} \label{sec:experiment}

\subsection{Experiments on Traveling Salesman Problem Instances}\label{subsec:exp:TSP}

Computational experiments were performed on 126 test instances with the known optimum from the library~\cite{Soler2008}, for which the number of vertices is varied from 64 to 315. These instances are generated from one series of instances of the  mixed general routing problem with turn penalties (see~\cite{Soler2008}) and characterized by asymmetric but similar-valued edge weights.

\begin{table}[!ht]
\caption{Comparison of results for algorithms: our GA (modes A, B, C) and $GA_{EAX}$ on ATSP instances}
\label{tab:MA+GA}
\centering
\begin{tabular}{|c|c|c|c|c|}
\hline
metric & GA (mode C) & GA (mode B) & GA (mode A) & $GA_{EAX}$ \\
\hline
$\Delta_{\mathrm{aver}}$ & 0.00048 & 0.00091 & 0.0017 & 0.00029 \\
\hline
$\Delta_{\mathrm{best}}$ & 0.00012 & 0.00041 & 0.0010 & 0.000024 \\
\hline
$C_{\mathrm{aver}}$ & 1229542.62 & 1229547.39 & 1229555.99 & 1229539.10 \\
\hline
$C_{\mathrm{best}}$ & 1229537.12 & 1229540.44 & 1229547.33 & 1229535.35 \\
\hline
\end{tabular}
\end{table}

All modes of our GA with two population sizes ($100$ and $300$) were executed 100 times under the same CPU time limit. The best results were achieved by GA versions with the population size equal to $300$. Let $C_{\mathrm{best}}$ and $C_{\mathrm{aver}}$ denote the best and average objective values over 100 runs on an instance. Let $\Delta_{\mathrm{aver}}$ and $\Delta_{\mathrm{best}}$ represent the percentage deviation of $C_{\mathrm{aver}}$ and $C_{\mathrm{best}}$ from the optimal objective value on an instance.
Experimental results in the context of $\Delta_{\mathrm{aver}}$, $\Delta_{\mathrm{best}}$, $C_{\mathrm{aver}}$ and $C_{\mathrm{best}}$ values averaged over all instances are presented in Table~\ref{tab:MA+GA}.  

Wilcoxon rank-sum test with significance level $\alpha=0.05$ was used to verify statistical significance of differences between methods in terms of $C_{\mathrm{best}}$ and $C_{\mathrm{aver}}$ values across all test instances. Mode C demonstrates better results than mode B, the difference between values of $C_{\mathrm{best}}$ and $C_{\mathrm{aver}}$ found by algorithms is statistically significant. 
The versions of the algorithm with restarts outperform the version without restarts, the differences in objective values are also statistically significant. So, the restarts
performed in the proposed  genetic algorithm allow to avoid localization of search and restore the population diversity, leading to better results
when the steady-state population management is used.

Our GA is also compared with one of the most competitive genetic algorithms for ATSP solution proposed in~\cite{NS12}. The algorithm from~\cite{NS12} ($GA_{EAX}$) uses an EAX crossover operator, it employs a population-based reproduction scheme (the size of population is 300) and uses 3-opt local optimization when building the initial population. This algorithm was implemented in C++ and tested on an Intel Xeon 2.93 GHz computer. Our algorithm  was implemented in Java and tested on an Intel Xeon E5420 2.5 GHz computer with 16 GB RAM.

$GA_{EAX}$ was also executed 100 times for each test instance with comparable computation time: our GA was allocated 1.465 times more time than the recorded computation time of $GA_{EAX}$. This multiplier was chosen based on CPU clock ratios and the fact that Java compiler is estimated to be 25\% slower than C++ compiler. The results are presented in Table~\ref{tab:MA+GA}.


The average deviation from optimum $\Delta_{\mathrm{aver}}$ for GA (mode C) is 0.00048, which is 1.5 times higher than for $GA_{EAX}$. However, GA (mode C) solved 82 instances with 100\% optimality frequency, compared to 68 instances for $GA_{EAX}$. No significant differences were found between average objective values $C_{\mathrm{aver}}$ for GA (mode C) and $GA_{EAX}$. The same holds for best found objective values $C_{\mathrm{best}}$. 


Previously in~\cite{EremeevKovalenko2020} we tested a similar GA (without postprocessing local optimization) on ATSP instances from TSPLIB library~\cite{Reinelt91}.
The collection includes 27 instances from different applications and
consists of three series ftv, ft, and rbg, and also individual
problems br17, p43, ry48p, kro124p. The dimensions of the instances
vary from 17 up to 443. 
A comparison of the modes~B and~C was done 
under the same CPU-time limit. The experiment was executed $1000$ times on each instance. The three versions of the algorithm were compared in terms of the frequencies of finding an optimum.

Mode~C demonstrated better results than mode~B on 12~instances, and the
difference between the frequencies of finding an optimum  was statistically significant in 5~cases (the significance test~\cite{BH1977} is used). 
Mode~B slightly outperformed mode~C on four test problems, but the difference was statistically significant only on instance kro124p. Modes B and C demonstrated statistically significantly better results in comparison with mode A on all instances from TSPLIB library. 

Moreover, the experimental evaluation from~\cite{EremeevKovalenko2020} has shown that the proposed algorithm (modes B and C)  yields results competitive or superior to those of other state-of-the-art  algorithms~\cite{BFM2004,XCYHSC2008,TWO14,NS12,RGG2016,BCK2001}. Note that most of the TSPLIB instances (except rbg series) exhibit more substantial edge weight differences than instances from~\cite{Soler2008}. Thus, we conclude that both our GA and $GA_{EAX}$ demonstrate competitive performance across problems with different structures.

\subsection{Experiments on Total Weighted Tardiness Problem Instances}\label{subsec:exp:TWTP}

We use the TWTP instances with the known optimal solutions from OR-Library.\footnote{http://people.brunel.ac.uk/~mastjjb/jeb/orlib/wtinfo.html} 
The number of jobs in the series of instances are 40, 50 and 100.
Each series contains 125 tests. Our  experiment war repeated on each instance for 50 times. 
The algorithm is set to stop when it obtains an optimal solution or reaches a maximum number of generations, equal to~6000  
(such condition is selected in order to compare our results to the ones form~\cite{AAS2003,CFLL2016,DLCX2017,WT2009}). 
We set $p_\mathrm{mut}=0.15$, $p_\mathrm{cross}=0.8$, and use the tournament selection with the tournament size $s = 5$.
Our experiment was carried out on a computer with Intel Core i3-10100F CPU 3.60~GHz, 16~Gb,
GA-LI~\cite{AAS2003} was  tested on a computer with Pentium~II 400MHz, 96Mb; 
PVNS~\cite{WT2009} was tested on a computer with Pentium~IV 3.0GHz, 512Mb;
m-VNS~\cite{CFLL2016} was tested on a computer with Intel Core 2.3GHz, 2Gb;
HEA-DS~\cite{DLCX2017} was tested on a computer with Xeon E5440 2.83GHz, 16Gb. 


\begin{table}
\begin{tabular}{ccccc}
Algorithm \ & \ Time, sec. \ & \ $n_{hit}$ \ & \ $n_{opt}$ \ & \ $\Delta_{aver}$ \\
\hline
GA-LI \ & \ 29.11  \ & \ --  \ & \ 100\% \ & \ 0\\ 

PVNS \ & \ 6.19  \ & \ 100\%  \ & \ 100\% \ & \ 0\\ 

m-VNS  \ & \ 0.908 \ & \ 100\%  \ & \ 100\% \ & \ --\\ 

GA (mode C) \ & \ 0.05  \ & \ 100\%  \ & \ 100\% \ & \ 0\\ 
GA (mode B) \ & \ 0.05  \ & \ 100\%  \ & \ 100\% \ & \ 0\\ 
GA (mode A) \ & \ 0.92  \ & \ 100\%  \ & \ 100\% \ & \ 0\\

HEA-DS \ & \ 0.003 \ & \ 100\%  \ & \ 100\% \ & \ 0\\
\hline
\end{tabular}

\caption{Results for TWTP series with $n=40$.}
\label{ex:40}
\end{table}

\begin{table}

\begin{tabular}{ccccc}
Algorithm \ & \ Time, sec. \ & \ $n_{hit}$ \ & \ $n_{opt}$ \ & \ $\Delta_{aver}$ \\
\hline
GA-LI \ & \ 41.02 \ & \ --  \ & \ 99.2\% \ & \ 0.000\\ 

PVNS \ & \ 12.15  \ & \ 100\%  \ & \ 100\% \ & \ 0\\ 

m-VNS \ & \ 1.62 \ & \ 100\%  \ & \ 100\% \ & \ --\\ 

GA (mode C) \ & \ 0.12  \ & \ 100\%  \ & \ 100\% \ & \ 0\\ 
GA (mode B) \ & \ 0.14  \ & \ 100\%  \ & \ 100\% \ & \ 0\\ 
GA (mode A) \ & \ 1.11  \ & \ 94\%  \ & \ 100\% \ & \ 0\\

HEA-DS \ & \ 0.006 \ & \ 100\%  \ & \ 100\% \ & \ 0\\

\hline
\end{tabular}
\caption{Results for TWTP series with $n=50$.}
\label{ex:50}
\end{table}

\begin{table}
\begin{tabular}{ccccc}
Algorithm \ & \ Time, sec. \ & \ $n_{hit}$ \ & \ $n_{opt}$ \ & \ $\Delta_{aver}$ \\
\hline
GA-LI \ & \ 118.97  \ & \ --  \ & \ 66.4\% \ & \ 0.02\\

PVNS \ & \ 183.47  \ & \ 100\%  \ & \ 100\% \ & \ 0\\

m-VNS \ & \ 5.845 \ & \ 99.84\%  \ & \ 100\% \ & \ --\\

GA (mode C) \ & \ 1.4 \ & \ 100\%  \ & \ 100\% \ & \ 0\\ 
GA (mode B) \ & \ 1.5 \ & \ 100\%  \ & \ 100\% \ & \ 0\\
GA (mode A) \ & \ 3.18 \ & \ 67\%  \ & \ 74\% \ & \ 0.017\\

HEA-DS \ & \ 0.032 \ & \ 100\%  \ & \ 100\% \ & \ 0\\
\hline
\end{tabular}
\caption{Results for TWTP series with $n=100$.}
\label{ex:100}
\end{table}

The results of the experiment are provided in Tables~\ref{ex:40},~\ref{ex:50} and~\ref{ex:100}. Here
$n_{hit}$ is the average percentage the cases when an optimal solution was found for an instance out of the given trial runs,
$n_{opt}$ is the percentage of instances, where the optimum is found, 
$\Delta_{aver}$ is the average relative deviation from the optimum, Time is the average CPU time for an instance. 

The  CPU time of mode C is lower than the CPU time of mode~B on series with $n=50,\ 100$ and the difference is statistically significant (the Wilcoxon rank-sum test with significance level $\alpha=0.05$ was used). GAs with restarts significantly outperformed the GA without restarts on all series in terms of CPU times and frequency of success in finding an optimum. 

We see that our algorithm with restarts in mode~C demonstrates competitive results, finding the best-known solutions in all cases within comparable time. 
Our algorithm shows slightly worse results than HEA-DS in terms of the  time for reaching of the best-known solutions but the main focus of the present paper was not a competition to other algorithms but experimental evaluation of the impact from usage of the new restart rule.


\section{Conclusions}\label{sec:concl}

A new restart rule is proposed for Genetic Algorithms, using the
Schnabel Census method, originally developed for statistical
estimation of size of animal populations. The performance of the
new restart rule is demonstrated on a steady-state GA, although
the specifics of this GA is not crucial for application of the proposed restart rule. Computational
experiments show a significant advantage of the GA with the new
restarting rule over the GA without restarts and the GA which is restarted
as soon as the current iteration number becomes twice the
iteration number of the currently best incumbent. 

Further improvements of the restart strategy are expected via
usage of less biased methods, developed for estimation of the
number of local optima and for estimation of the abundance of
populations, see e.g.~\cite{ER,Hernando}. The experimental research in future
might also address the usefulness of the proposed restart rule for other
types of evolutionary algorithms and other optimization problems.
The Schnabel Census method may be also applicable for a dynamical control  of the variation operators in GAs, if instead of restarting the GA, it will change the probability distribution of mutation and/or crossover.

\section*{Funding}
The work is supported by the Mathematical Center in Akademgorodok under the agreement \textnumero~075-15-2025-349 with the Ministry of Science and Higher Education of the Russian Federation.

\section*{Code availability }
Java code of the restart rule and GAs from Section~\ref{sec:GAs_on_permut} are available at \\
https://gitlab.com/YuliaKovalenko$-$gl/schnabel$\_$census/

\end{document}